\documentclass{article}

\PassOptionsToPackage{numbers, compress}{natbib}

\usepackage[preprint]{airov24}

\usepackage[utf8]{inputenc} 
\usepackage[T1]{fontenc}    
\usepackage{hyperref}       
\usepackage{url}            
\usepackage{booktabs}       
\usepackage{amsfonts}       
\usepackage{nicefrac}       
\usepackage{microtype}      
\usepackage{graphicx}
\usepackage{multirow}
\usepackage[table,xcdraw]{xcolor}

\title{Prompt-Time Symbolic Knowledge Capture with Large Language Models}

\author{
  Tolga Çöplü, Arto Bendiken, Andrii Skomorokhov, Eduard Bateiko, \AND Stephen Cobb, Joshua J. Bouw\\
  Haltia, Inc. \\
  \texttt{\{tolga, arto, andrii, eduard, steve, joshua\}@haltia.ai} \\
}

\begin{document}

\maketitle

\begin{abstract}
Augmenting large language models (LLMs) with user-specific knowledge is crucial for real-world applications, such as personal AI assistants. However, LLMs inherently lack mechanisms for prompt-driven knowledge capture. This paper investigates utilizing the existing LLM capabilities to enable prompt-driven knowledge capture, with a particular emphasis on knowledge graphs. We address this challenge by focusing on prompt-to-triple (P2T) generation. We explore three methods: zero-shot prompting, few-shot prompting, and fine-tuning, and then assess their performance via a specialized synthetic dataset. Our code and datasets are publicly available at \href{https://github.com/HaltiaAI/paper-PTSKC}{https://github.com/HaltiaAI/paper-PTSKC}
\end{abstract}

\section{Introduction}
Large language models (LLMs) are transforming human-machine interaction with their adeptness in carrying out conversations. However, despite their proficiency in responding to queries, LLMs cannot be considered good listeners. Their limitation lies in their inability to learn from user-provided information. To utilize any data received from users beyond their context window, LLMs require support from an external system, a significant gap in their interactive capabilities.

Primarily, LLMs capture knowledge during their training phase. This phase, which enables the implicit encoding of knowledge into the model's parameters, is deemed efficient in terms of data compression. However, the substantial computational power, time, and cost required for training, particularly in the pre-training stage, render it impractical for prompt-driven continuous learning. LLMs are unable to capture knowledge obtained from users or through external integrations/plugins, which presents significant challenges for many AI applications. For instance, the ability of AI assistants to capture and utilize personal information in future interactions is crucial. This limitation is currently being addressed through various Retrieval-Augmented Generation (RAG) approaches. Within this realm, knowledge graphs are distinguished by their clear structures, symbolic representations, and capacity for factual reasoning, making their integration with LLMs a vibrant area of ongoing research \citep{pan_large_2023,pan_unifying_2023}.

In this paper, we focus on the building blocks of prompt-driven symbolic knowledge capture. We investigate the extraction of prompts in subject-predicate-object triples\footnote{https://www.w3.org/TR/rdf12-concepts/} for a predefined context (relation) through in-context learning and fine-tuning approaches. Utilizing a specially designed dataset, we aim to assess the efficacy of these methods, highlighting their strong points and identifying areas for enhancement.

The structure of this paper is as follows: Section 2 introduces the proposed in-context learning and fine-tuning approaches by providing examples. Section 3 describes the experimental setup by presenting the development framework, the language model selection, and the dataset creation process.  Section 4 outlines our test results and their interpretations. Finally, Section 5 concludes the paper and suggests future directions.

\section{Prompt-to-triple generation}
Triples, composed of \textit{('subject', 'predicate', 'object')}, are considered a universal data model thanks to their inherent simplicity and versatility. This format reflects the fundamental structure of human language and cognition, capturing the essence of any asserted statement or fact. Each triple represents a distinct atom of knowledge, with the subject and object identifying entities and the predicate describing their relationship. In our study, we have chosen triples as our data model for these characteristics. Furthermore, for ease of presentation, we employ an informal free-form triple format, which allows for greater flexibility in our discussions and examples.

Generating triples based on a predefined context from user prompts can be viewed as a specific aspect of the broader text-to-graph (T2G) generation problem \citep{guo_cyclegt_2020, xu_infinity_2022}. This perspective led us to define our research problem as prompt-to-triple (P2T) generation. P2T generation entails extracting 'subject' and 'object' terms from user prompts that correspond with a 'predicate' drawn from a restricted vocabulary. This vocabulary consists of predefined, user-specific relations such as birthdays, anniversaries, locations, and events. A key aspect is ensuring that the 'predicate' term of the generated triple accurately reflects the relevant predefined relation. For example, from the prompt 'I was born in 1979', our goal is to generate the triple \textit{('I', 'birthday', '1979')}, aligning with the 'birthday' relation.

In our research, we began by pinpointing some relevant predefined relations (aka a restricted vocabulary for the ‘predicate’ term). Following on this, we developed the requisite training and test datasets essential for addressing the problem. Building on this groundwork, we have formulated the following three methodologies to effectively tackle the P2T generation challenge

\subsection{P2T zero-shot prompting}
Zero-shot prompting is an in-context learning technique that enables LLMs to apply their inherent knowledge to complete tasks without additional training. This approach is relevant in the T2G generation domain, especially in the multi-turn question answering form, as noted in \citep{li_entity-relation_2019}. However, the P2T generation task requires tailored zero-shot prompts due to specific predefined relations. Our approach has evolved through a series of iterative developments and tests. The critical aspects of zero-shot prompting, as explored in our research, include:

\begin{figure}[ht]
  \centering
  \includegraphics[scale=0.70]{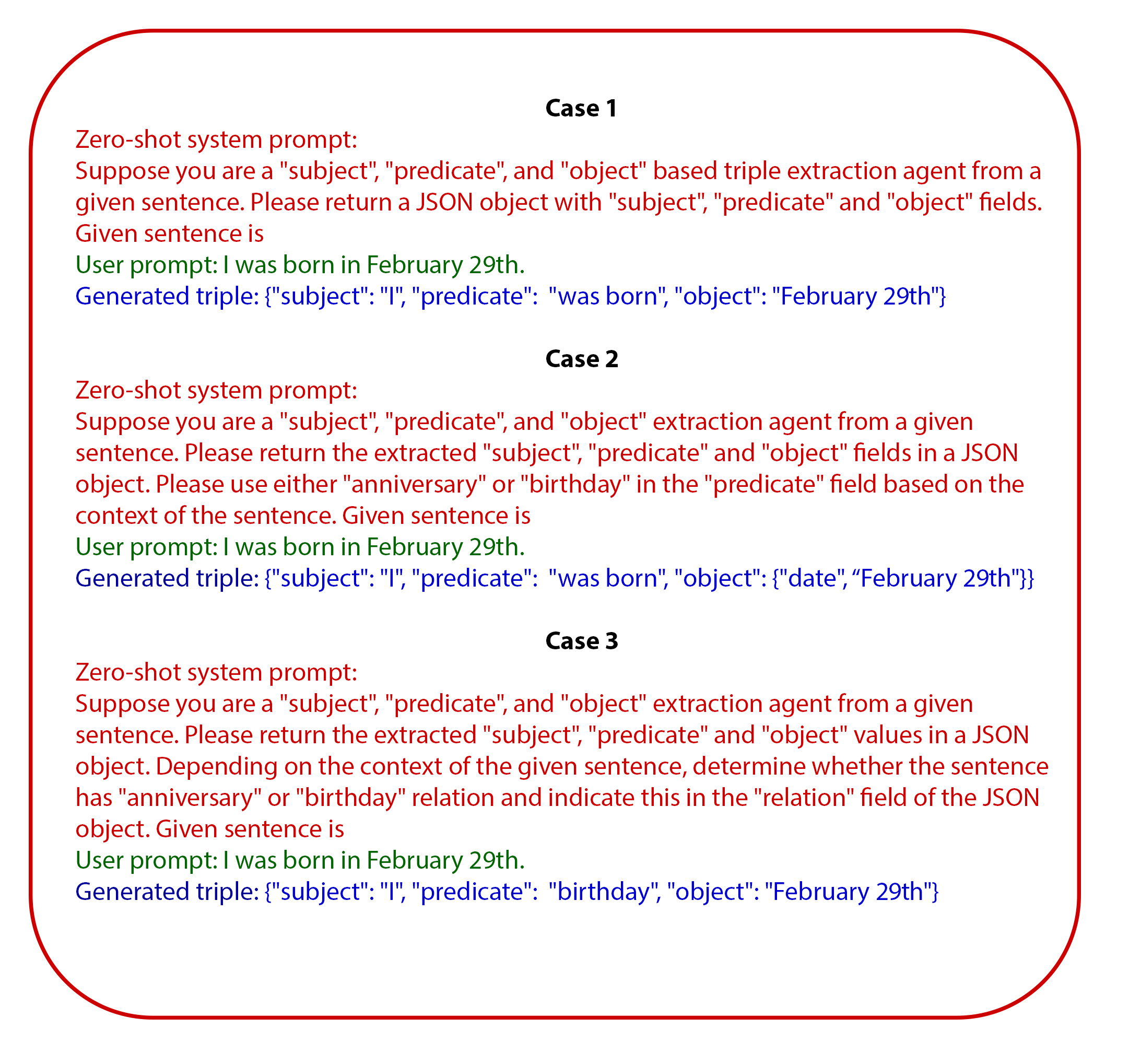}
  \caption{Cases of zero-shot prompting to demonstrate the evaluation.}
  \label{fig:fig1}
\end{figure}

\begin{figure}[ht]
  \centering
  \includegraphics[scale=0.7]{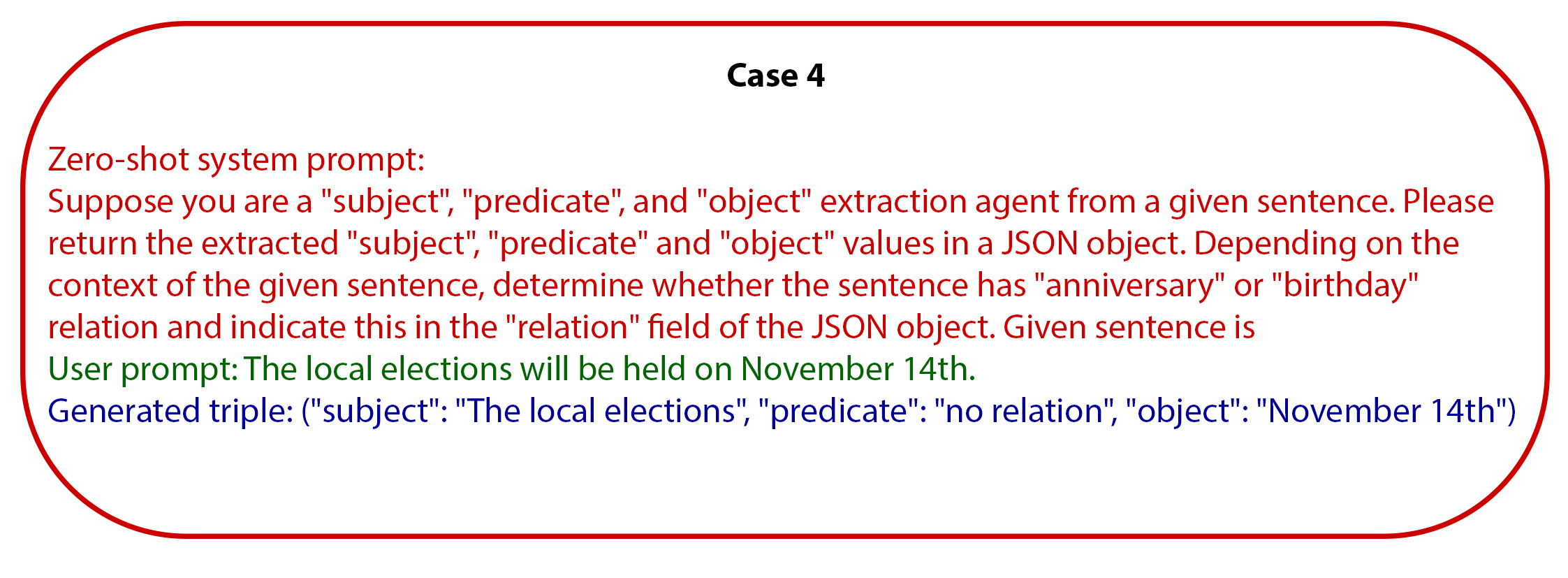}
  \caption{Zero-shot prompting case for out-of-context input}
  \label{fig:fig2}
\end{figure}

\begin{itemize}
\item For the prompt context to match the predefined relations, the set of predefined relations must be present in the system prompt. This leads to scalability issues, as both the prompt size and processing time vary with the size of the relation set.
\item Due to pre-training biases, LLMs often default to assigning the sentence's verb to the 'predicate' term. This can result in incorrect triples when the verb doesn't match the predefined relation, despite explicit instructions for relation matching. To improve accuracy, we introduced an extra term for relation matching, which is then incorporated into the ‘predicate’ term via post-processing. Relevant cases are presented in Figure.\ref{fig:fig1}. Case 1 demonstrates that, in the absence of explicit instructions, the LLM generates the ‘predicate’ by utilizing the verb from the provided sentence, aligning with expectations. In Case 2, despite clear instructions for relation matching, the selection of either 'anniversary' or 'birthday' as the predicate is suppressed. Case 3 resolves this by using a new 'relation' term specifically for matching 'birthday' or 'anniversary’.
\item LLMs can recognize that a sentence context falls outside the predefined relation set, even without explicit instruction in the zero-shot prompt. The LLM even adds a justification in the response. An instance of this scenario is depicted in Figure.\ref{fig:fig2}.
\end{itemize}

\subsection{P2T few-shot prompting}
Few-shot prompting, an in-context learning technique, provides an LLM with a small set of examples to enhance its task understanding before generating a response \citep{brown_language_2020}. This method stands apart from zero-shot prompting, which requires no examples, and fine-tuning, which demands extensive training data. Few-shot prompting seeks a balance by offering sufficient context for efficient model operation. Although there are criticisms in the literature \citep{ma_large_2023,zhu_llms_2023} regarding the efficiency of few-shot prompting, in our study, we aimed to evaluate this method using our own dataset. Mirroring the approach used in zero-shot prompting, we adopted an iterative development and testing cycle for few-shot prompting. The significant aspects of few-shot prompting are outlined below:

\begin{itemize}
\item In a few-shot prompting, providing an example for every predefined relation is necessary. This requirement, similar to zero-shot prompting, leads to scalability challenges.
\item The variety of examples has a direct impact on performance. To effectively match the 'birthday' relation, examples must cover various sentence structures, such as “I was born in 1979” and “My birthday is in November”. Relevant cases are presented in Figure.\ref{fig:fig3}.

\begin{figure}[ht]
  \centering
  \includegraphics[scale=0.75]{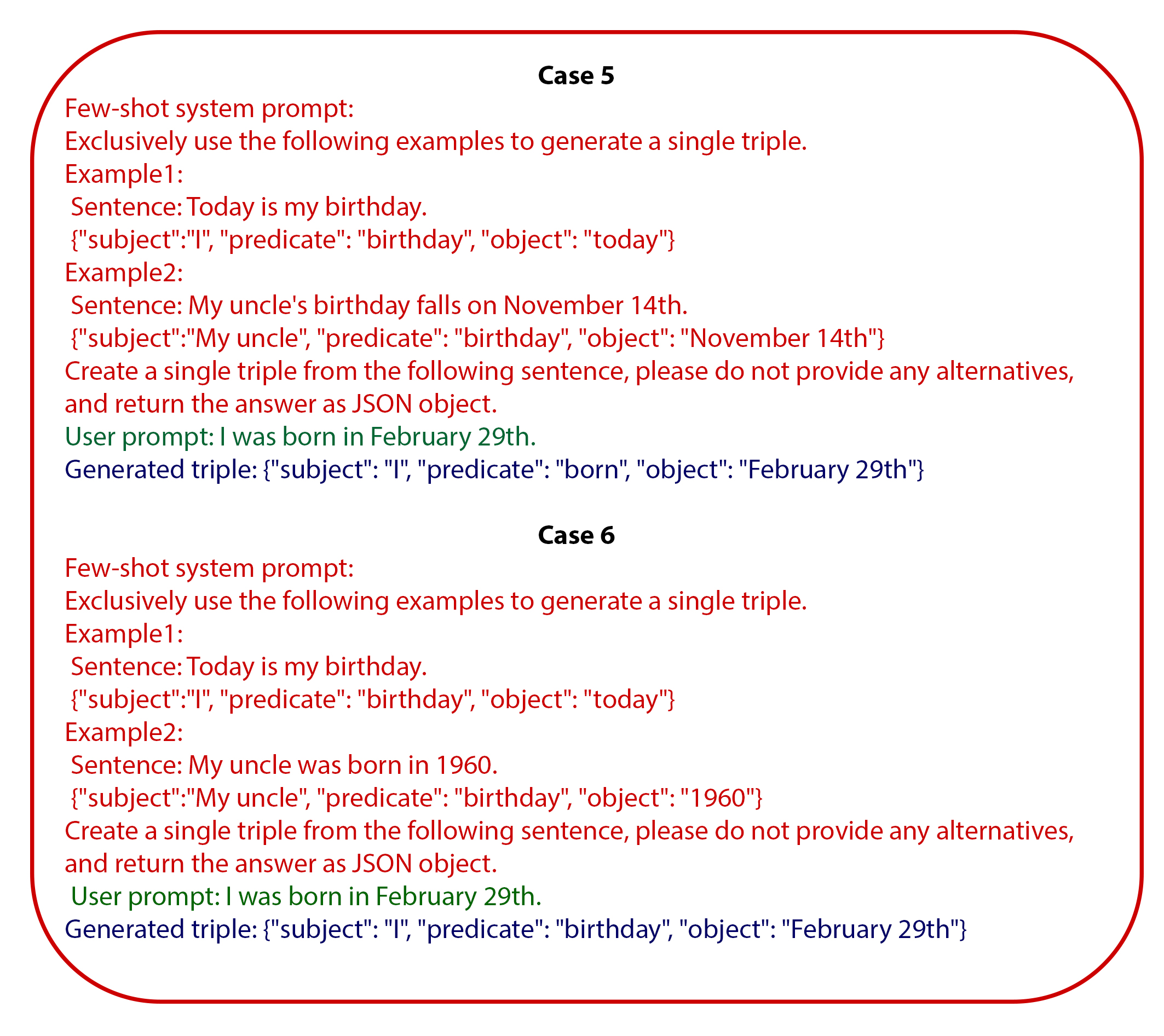}
  \caption{The impact of example diversity is presented with two different few-shot prompts}
  \label{fig:fig3}
\end{figure}

\item When the LLM encounters a sentence context outside the predefined relation set, it relies on its implicit knowledge to perform triple extraction, due to the absence of a corresponding example in the few-shot prompt. Figure.\ref{fig:fig4} represents a case of this particular scenario.
\end{itemize}

\begin{figure}[ht]
  \centering
  \includegraphics[scale=0.72]{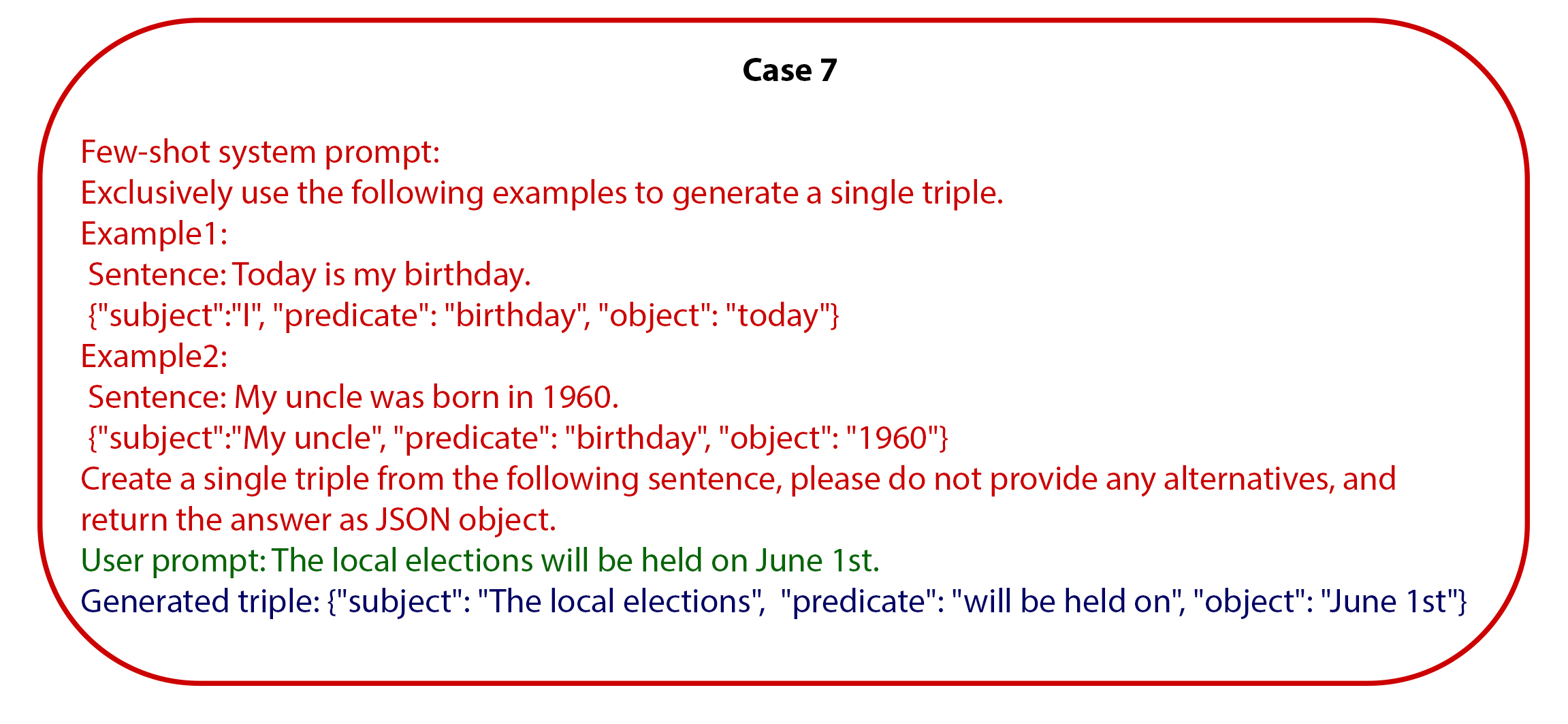}
  \caption{Few-shot prompting example for out of context input}
  \label{fig:fig4}
\end{figure}

\subsection{P2T generation using a fine-tuned LLM}
\begin{figure}[ht]
  \centering
  \includegraphics[scale=0.75]{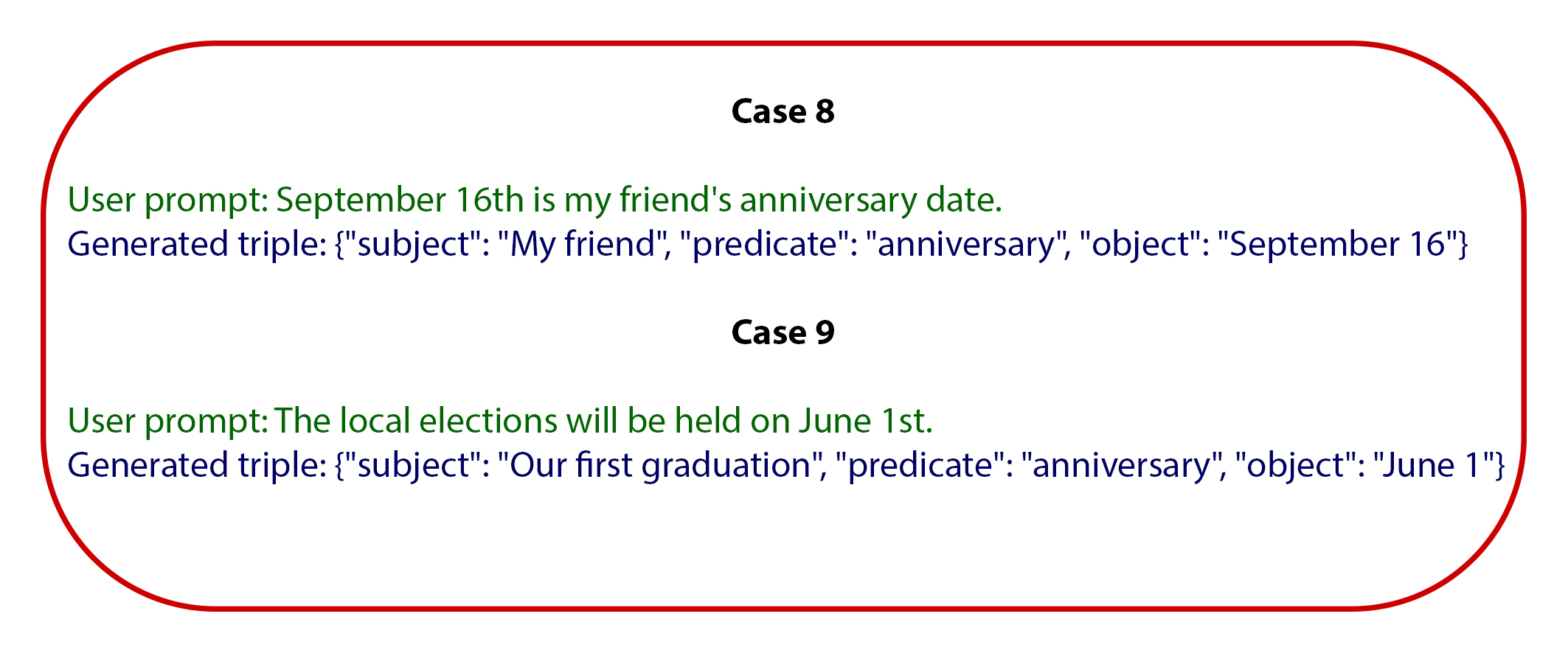}
  \caption{Fine-tuning examples for in-context and out-of-context prompts}
  \label{fig:fig5}
\end{figure}

Fine-tuning is a process where a pre-trained LLM is further trained on a specific dataset. This technique adjusts the model's parameters to make it more adept at a particular task, such as P2T generation in our case. The following points highlight the key aspects of P2T fine-tuning:

\begin{itemize}
\item The training dataset is critical to the success of the fine-tuning process. It provides a targeted environment with specific examples for relation matching and extracting subjects and objects, requiring a diverse array of examples to address all potential issues. Details regarding the dataset and its generation are presented in Section \ref{dataset_ref}.
\item For each predefined relation, the training dataset must contain varied examples. This requirement does not lead to scalability issues as in zero-shot or few-shot prompting, but an enlarged training set might increase the risk of performance degradation in other tasks for the LLM.
\item When the LLM encounters a sentence context not present in the predefined relation set, it resorts to its implicit knowledge for triple extraction, similar to the few-shot prompting scenario. Figure.\ref{fig:fig5} presents cases for relations both included and excluded in the training set.
\end{itemize}

\section{Experimental setup}
This section explores the components of our experimental setup.

\subsection{Development framework}
The methods suggested in this paper have been implemented using the Apple MLX framework \citep{mlx2023}. MLX is a specialized array framework designed for machine learning applications, akin to NumPy, PyTorch, or Jax, with the distinction of being exclusive to Apple silicon.

P2T fine-tuning has been conducted using the parameter-efficient QLoRA approach \citep{dettmers_qlora_2023} on our custom dataset, comprising randomly selected, non-overlapping sets of 1,000 training, 200 validation, and 200 test samples. The fundamental QLoRA parameters used are as follows:

\begin{itemize}
\item Optimizer: Adam
\item Learning rate: \( 1x10^{-5} \)
\item Number of layers to fine-tune: 16
\item Minibatch size: 4
\item Iterations: 1,000
\end{itemize}

\subsection{LLM}
The methods we have developed here do not have a structural dependency on a particular underlying foundation model.. The key factors guiding our LLM selection were its proven effectiveness across diverse domains in community benchmarks and its prevalence in the field. Owing to its performance in the Hugging Face Leadership benchmark \citep{open-llm-leaderboard} and its robust ecosystem, the Mistral-7B-Instruct-v0.2 \citep{jiang_mistral_2023}, based on the Llama 2 \citep{touvron2023llama} architecture, was selected for our research. We ran all examples, tests, and benchmarks on a 4-bit quantized version of this model.

\subsection{Dataset} 
\label{dataset_ref}
In the field of natural language processing (NLP), the creation of robust and diverse datasets is crucial for training and evaluating LLMs, especially for tasks such as knowledge extraction, which involves identifying structured information from unstructured text. Aligning with these needs, we have created a synthetic dataset focused on 'birthday' and 'anniversary' content for P2T generation study, adhering to the detailed stages described below.

Our synthetic dataset creation process consists of three distinct stages:
\begin{enumerate}
\item We engaged native speakers to write templates for 86 user prompts and model responses. This initial step ensures the dataset's foundational accuracy and contextual relevance.
\item We leveraged Python's capabilities, particularly a random date generator, to expand this dataset from 86 to 860 prompt-response pairs.
\item The final stage of our dataset development involved using an LLM, specifically Llama-2-7b-chat-hf, to paraphrase each prompt from the previous dataset five times. This resulted in a new dataset of 4300 prompt-response pairs. Paraphrasing is a critical step as it introduces linguistic variations and nuances, thereby enriching the dataset and making it more representative of natural language variability \citep{feng_survey_2021, li_data_2022}. This approach is supported by recent studies which highlight the importance of paraphrasing in dataset creation for NLP tasks, as it significantly contributes to model generalizability and understanding of diverse linguistic expressions.
\end{enumerate}

\section{Performance evaluation}
We have evaluated the proposed methods using a non-overlapping test dataset, comprising 128 'birthday' and 72 'anniversary' relations. The outputs generated during the evaluation were compared with manually crafted ground-truths. The comparisons were conducted in two distinct manners: ‘relation-based’ and ‘triple-based’.

\begin{itemize}
\item \textit{\textbf{Relation-based comparison:}} This approach focuses solely on the 'predicate' term. In this scenario, equality between the test result and the ground-truth value is reported as a True Positive (TP). 'Predicate' values falling outside the predefined relation set are reported as False Negatives (FN), while those within the set but differing from the ground-truth are reported as False Positives (FP).
\item \textbf{\textit{Triple-based comparison:}} This method involves comparing all terms of the generated triple. The comparison of the 'predicate' term follows the same approach as the relation-based method. However, the 'subject' and 'object' values are compared based on a relationship of inclusion rather than direct equality. For example, a generated triple \textit{('I', 'birthday', 'November 14th')} compared with the ground truth \textit{('I me', 'birthday', 'November 14')} is classified as TP.
\end{itemize}

The macro precision, recall, and f1-score calculated for both the relation-based and entity-based approaches are presented in Table \ref{tab:table1}.

As indicated in Table \ref{tab:table1}, the recall for relation-based evaluation is impeccable across all methods. We believe this outcome is associated with the tests being conducted on only two relations. When considering precision and f1-score, it is observed that both zero-shot prompting and fine-tuning methods emerge as prominent. We assess that the clear guidance provided by the zero-shot prompt, instructing the LLM to select one of the predefined relations, plays a significant role in its superior performance compared to few-shot prompting. The fact that the fine-tuning method yields significantly good results clearly demonstrates its success in learning straightforward tasks.

Upon examining the triple segment in Table \ref{tab:table1}, we observe that despite differing precision and recall performance, zero-shot prompting and few-shot prompting methods exhibit similar f1-scores. In this segment, fine-tuning has demonstrated superior performance compared to other methods. These outcomes motivate us to focus on fine-tuning and conduct more comprehensive studies on it.

\begin{table}[ht]
\centering
\caption{Relation and triple generation performance based on macro precision, recall and f1-score.}
\label{tab:table1}
\begin{tabular}{|l|ccc|ccc|}
\hline
                    & \multicolumn{3}{c|}{\cellcolor[HTML]{EFEFEF}Relation}          & \multicolumn{3}{c|}{\cellcolor[HTML]{EFEFEF}Triple}                \\ \cline{2-7} 
\multirow{-2}{*}{} &
  \multicolumn{1}{l|}{\cellcolor[HTML]{FFFFC7}Precision} &
  \multicolumn{1}{l|}{\cellcolor[HTML]{FFFFC7}Recall} &
  \multicolumn{1}{l|}{\cellcolor[HTML]{FFFFC7}F1-score} &
  \multicolumn{1}{l|}{\cellcolor[HTML]{FFFFC7}Precision} &
  \multicolumn{1}{l|}{\cellcolor[HTML]{FFFFC7}Recall} &
  \multicolumn{1}{l|}{\cellcolor[HTML]{FFFFC7}F1-score} \\ \hline
zero-shot prompting & \multicolumn{1}{c|}{0.815} & \multicolumn{1}{c|}{1.0} & 0.8981 & \multicolumn{1}{c|}{0.6636} & \multicolumn{1}{c|}{0.4479} & 0.5348 \\ \hline
few-shot prompting  & \multicolumn{1}{c|}{0.49}  & \multicolumn{1}{c|}{1.0} & 0.6577 & \multicolumn{1}{c|}{0.3855} & \multicolumn{1}{c|}{0.6531} & 0.4848 \\ \hline
fine-tuning         & \multicolumn{1}{c|}{1.0}   & \multicolumn{1}{c|}{1.0} & 1.0    & \multicolumn{1}{c|}{1.0}    & \multicolumn{1}{c|}{0.96}   & 0.9796 \\ \hline
\end{tabular}
\end{table}

\section{Conclusion}
In this paper, we initially discussed prompt-driven symbolic knowledge capture and its significance in the LLM domain. We then projected the prompt-driven symbolic knowledge capture problem into prompt-to-triple (P2T) generation, which involves generating triples based on predefined relations from user prompts. To address P2T, we developed new approaches using fundamental LLM techniques, including in-context learning and fine-tuning. We concluded our work with performance evaluations of these proposed methods.

Our findings indicate that fine-tuning is particularly sensitive in addressing P2T. In our future work, we aim to refine the fine-tuning approach and comprehensively examine its impact on the overall performance of the model across across various scenarios.

Please see the corresponding GitHub repository at \href{https://github.com/HaltiaAI/paper-PTSKC}{https://github.com/HaltiaAI/paper-PTSKC}


\bibliographystyle{unsrtnat}

\bibliography{references}  

\end{document}